%% file: neurips_2025.tex
\documentclass{article}

\PassOptionsToPackage{numbers, compress}{natbib}

\usepackage{multirow} 
\usepackage{multicol}

\usepackage[table]{xcolor} 

\usepackage[most]{tcolorbox} 

\usepackage{CJKutf8}
\usepackage[preprint]{neurips_2025}

\usepackage[utf8]{inputenc} 
\usepackage[T1]{fontenc}    
\usepackage{hyperref}       
\usepackage{url}            
\usepackage{booktabs}       
\usepackage{amsfonts}       
\usepackage{nicefrac}       
\usepackage{microtype}      
\usepackage{xcolor}         
\usepackage{graphicx}
\usepackage{booktabs}
\usepackage{enumitem}

\usepackage{array}   
\usepackage{wrapfig}
\usepackage{times}
\usepackage{latexsym}
\usepackage{pifont}
\usepackage[T1]{fontenc}

\usepackage{amssymb}

\definecolor{kellygreen}{rgb}{0.3, 0.73, 0.09}
\definecolor{alizarin}{rgb}{0.82, 0.1, 0.26}
\usepackage{amsmath}
\definecolor{lightpurple}{RGB}{147, 112, 219} 
\definecolor{lightgray}{RGB}{211, 211, 211} 
\definecolor{lightorange}{RGB}{255, 200, 120} 
\definecolor{lightred}{RGB}{255, 182, 193} 

\definecolor{cc}{rgb}{0.1, 0.2, 0.8} 
\definecolor{bb}{rgb}{0.3, 0.4, 0.85}

\definecolor{lightblue}{HTML}{ebf3f8}

\definecolor{mediumblue}{HTML}{d7e8f2}

\definecolor{deepblue}{HTML}{c8dfed}

\newcommand{\cc}[1]{\textcolor{cc}{\textbf{#1}}}

\title{SuperWriter: Reflection-Driven Long-Form Generation with Large Language Models}

\title{SuperWriter: Reflection-Driven Long-Form Generation with Large Language Models}

\author{%
Yuhao Wu$^{1}$\thanks{Equal contribution.},\hspace{0.5em}%
Yushi Bai$^{2}$\footnotemark[1],\hspace{0.5em}%
Zhiqiang Hu$^{1}$,\hspace{0.5em}%
Juanzi Li$^{2}$,\hspace{0.5em}%
Roy Ka-Wei Lee$^{1}$\\[1ex]
$^{1}$Singapore University of Technology and Design, Singapore\\
$^{2}$Tsinghua University, Beijing, China
}

\begin{document}

\maketitle

\input{1_abstract}

\input{2_introduction}

\input{4_1_Agent}

\input{4_2_method}

\input{5_experiment}

\input{3_related_work}

\input{6_conclusions}

\bibliographystyle{plainnat}
\bibliography{neurips_2025}


\appendix
\input{7_APPENDIX}


\newpage

\end{document}

%% file: 1_abstract.tex
\begin{abstract}

Long-form text generation remains a significant challenge for large language models (LLMs), particularly in maintaining coherence, ensuring logical consistency, and preserving text quality as sequence length increases. To address these limitations, we propose \textit{SuperWriter}-Agent, an agent-based framework designed to enhance the quality and consistency of long-form text generation. \textit{SuperWriter}-Agent introduces explicit structured thinking—through planning and refinement stages—into the generation pipeline, guiding the model to follow a more deliberate and cognitively grounded process akin to that of a professional writer.
Based on this framework, we construct a supervised fine-tuning dataset to train a 7B \textit{SuperWriter}-LM.
We further develop a hierarchical Direct Preference Optimization (DPO) procedure that uses Monte Carlo Tree Search (MCTS) to propagate final quality assessments and optimize each generation step accordingly. Empirical results across diverse benchmarks demonstrate that \textit{SuperWriter}-LM achieves state-of-the-art performance, surpassing even larger-scale baseline models in both automatic evaluation and human evaluation. Furthermore, comprehensive ablation studies demonstrate the effectiveness of hierarchical DPO and underscore the value of incorporating structured thinking steps to improve the quality of long-form text generation. Our code \& models are at: \url{https://github.com/mozhu621/SuperWriter}.
\end{abstract}

%% file: 2_introduction.tex
\begin{figure}[htbp]
    \centering
    \includegraphics[width=0.7\linewidth]{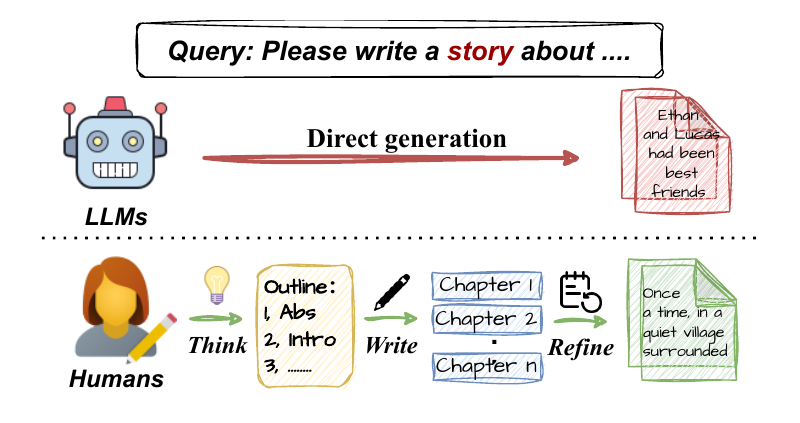}
    \caption{Current LLMs directly generate long text in a single pass, while human writers follow an iterative process of \textit{thinking}, \textit{outlining}, \textit{writing}, and \textit{refining} to ensure coherence and quality.}
    \label{fig:intro}
\end{figure}

\section{Introduction}

Effective long-form text generation remains a fundamental challenge for advancing large language models (LLMs)~\cite{yao2019planandwritebetterautomaticstorytelling,schmidgall2025agentlaboratoryusingllm,2024autosurvey}. Unlike short-form generation tasks, where LLMs have demonstrated remarkable success, generating extended sequences often lead to degraded coherence and logical consistency as the length grows.

Previous studies on long-form writing~\cite{bai2024longwriter,pham2024suri} typically employ LLMs to generate text in a single pass without structured intermediate thinking or explicit planning. Consequently, the generated texts often appear plausible initially but lack sustained coherence and can exhibit logical contradictions over longer spans~\cite{goldfarbtarrant2019planwritereviseinteractive,hua2020pairplanningiterativerefinement,rasheed2025largelanguagemodelscode,minaee2024largelanguagemodelssurvey,wu2025shiftinglongcontextllmsresearch}.

In practice, human writers rarely produce complex or lengthy documents in a single uninterrupted attempt. Instead, they commonly utilize iterative processes, including outlining, drafting, reviewing, and revising their writing to maintain coherence and logical consistency~\cite{flower1981cognitive,becker2006review,gollins2016framework}, as shown in Fig \ref{fig:intro}. Moreover, such writing processes are often accompanied by extensive cognitive activities, such as planning, foreshadowing, and strategic structuring—especially in tasks like writing detective novels or academic papers~\cite{flower1981cognitive}. Building on this observation, we propose \textit{SuperWriter}-agent, a novel agent pipeline explicitly designed to embed structured thinking paradigms within long-form text generation. By incorporating explicit intermediate cognitive steps, \textit{SuperWriter}-agent simulates the human writing process through a coordinated agent-based framework. 

Formally, \textit{SuperWriter}-Agent generates training data through a structured three-stage framework: Planning $\rightarrow$ Writing $\rightarrow$ Refining. In the Planning stage, two agents collaboratively reflect and outline key arguments, decompose complex ideas, and establish logical connections. During the Writing stage, each paragraph is composed based on the structured plan, incorporating explicit thinking steps into the content. In the Refining stage, the text is carefully reviewed and revised to ensure clarity and structural integrity. Unlike previous SFT datasets for long-form generation~\cite{bai2024longwriter,pham2024suri}, our approach explicitly segments the data into these three stages, each enriched with thinking-oriented supervision. To further guide the \textit{SuperWriter}-LM's writing capabilities, we apply hierarchical Direct Preference Optimization by comparing pairs of finalized response~\cite{rafailov2023direct} and propagate the feedback to each step through a Monte Carlo Tree Search (MCTS). Through the integration of structured thinking signals and hierarchical feedback, \textit{SuperWriter}-LM demonstrates notable improvements in fluency, coherence, and overall textual quality.

Finally, we validate the effectiveness of our trained \textit{SuperWriter}-LM through both human and automatic evaluations on the WritingBench benchmark~\cite{wu2025writingbenchcomprehensivebenchmarkgenerative}. Extensive ablation studies further underscore the critical role of structured thinking data and hierarchical Direct Preference Optimization (DPO) in enabling high-quality text generation. Our contributions are summarized as follows:

\begin{itemize}[leftmargin=*,itemsep=0pt,topsep=0pt]
    \item We introduce \textit{SuperWriter}-agent, an agent-based framework with a structured Planning--Writing--Refining workflow that simulates human writing processes by injecting intermediate thinking steps into long-form text generation.
    \item We construct a thinking-supervised, stage-segmented dataset and train the \textit{SuperWriter}-LM on it, then apply hierarchical DPO over final output comparisons to optimize each generation step.
    \item We empirically demonstrate the effectiveness of our approach, showing substantial improvements in fluency, coherence, and logical consistency compared to existing methods, as evaluated on \textit{WritingBench}~\cite{wu2025writingbenchcomprehensivebenchmarkgenerative}.
\end{itemize}

%% file: 4_1_Agent.tex
\section{\textit{SuperWriter}-Agent}

While current LLM training corpora provide abundant supervision data reflecting intermediate ``\textit{thinking}'' processes—such as mathematical reasoning and code generation~\cite{deepseekai2025deepseekr1incentivizingreasoningcapability}; however, they contain remarkably little data of this kind for writing tasks~\cite{cerebras2023slimpajama}. Most pretraining data~\cite{parmar2024data,weber2024redpajama} consists of finished texts like articles and books, which largely omit the underlying process of planning, structuring, and thinking involved in writing. However, writing—particularly long-form writing—inherently presents a complex cognitive task, where explicit thinking steps are crucial for maintaining coherence, logical flow, and structural consistency. 

\begin{figure*}[t!]
    \centering
    \includegraphics[width=\linewidth]{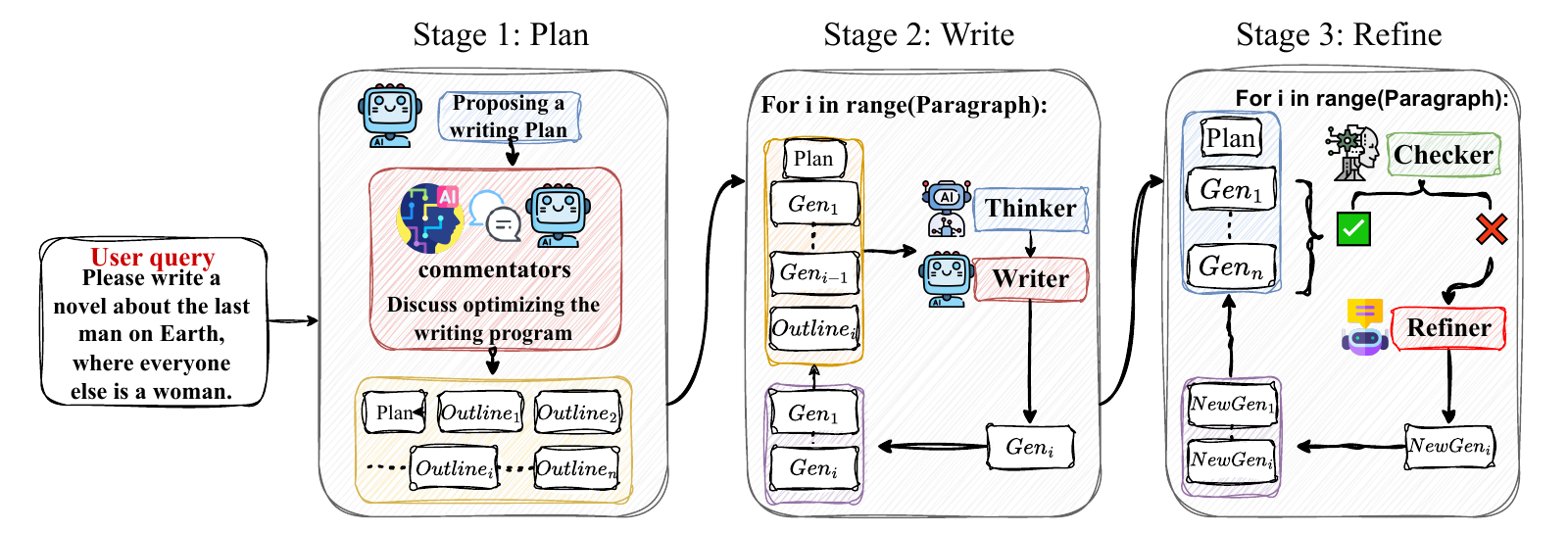}
    \caption{This figure illustrates a three-stage agent framework for long-form generation. In Stage 1 (Plan), the framework proposes a structured writing plan through discussions between AI commentators and a writer. In Stage 2 (Write), the text is incrementally generated using a thinker-writer collaboration, and in Stage 3 (Refine), a checker and refiner iteratively improve the generated text to enhance coherence and quality.}
    \label{fig:agent}
\end{figure*}

To address this gap, we propose the \textit{SuperWriter}-agent (illustrated in Figure~\ref{fig:agent}), a framework designed to generate high-quality, thought-enriched supervised fine-tuning data for writing. The \textit{SuperWriter}-agent enables structured content generation through three coordinated stages: careful planning, targeted paragraph-level writing, and iterative refinement. This process explicitly embeds intermediate thinking signals into the writing pipeline, thereby enhancing the fluency, coherence, and narrative consistency of the generated text.

\subsection{Stage 1: Plan}
\noindent

Inspired by the widely adopted pedagogical technique in writing education known as the Story Workshop\footnote{The Story Workshop is a writing instruction method developed at Columbia College Chicago. It emphasizes oral storytelling, collaborative discussion, and reflective dialogue to help writers generate ideas, structure narratives, and refine language through an interactive process.}~\cite{Schultz1977,tatiana2021collaborative,sheflett1973story}, Stage 1 of \textit{SuperWriter}-agent begins with oral narration and iterative dialogue aimed at distilling and expanding initial ideas. In practice, this planning stage guides discussion agents to articulate core themes, central arguments, character background settings (for genres like fiction), and paragraph-level content structures—collectively forming the Background component. By systematically allocating word counts and associating key ideas with specific paragraph units, this step builds a comprehensive and detailed outline for downstream writing. This structured process significantly enhances the overall coherence and organization of the text. With such a framework in place, discussion agents can strategically develop and refine their ideas, resulting in more focused, coherent, and well-developed written outputs. Appendix~\ref{Stage-1:prompt} provides the detailed prompt for the planning stage.

\subsection{Stage 2: Write}

Motivated by recent advancements in reasoning-oriented LLMs—notably their remarkable scalability during inference in tasks such as mathematical reasoning and code generation—we build upon paradigms exemplified by reasoning models like OpenAI o1 and DeepSeek-R1.~\cite{o1-preview,deepseekai2025deepseekr1incentivizingreasoningcapability}. These models typically engage in an explicit reasoning and planning phase prior to generating final responses, which has proven highly effective in enhancing output quality. Following this paradigm, we propose a two-stage generation framework that simulates the process of “thinking before writing,” aiming to produce structurally coherent and logically consistent paragraphs. Appendix~\ref{Stage-2:prompt} provides the detailed prompt for the actual paragraph write stage.

\begin{itemize}[leftmargin=*,itemsep=0pt,topsep=0pt] \item \textbf{Thinker Step:} In this initial phase, the model refrains from generating surface-level text. Instead, it identifies and organizes key ideas, thematic elements, and logical structures relevant to the paragraph. This explicit reasoning process provides a clear directional scaffold for subsequent text generation.

\item \textbf{Writer Step:} Building on the structured outline from the Thinker stage and incorporating the preceding paragraph (i.e., the  $(n-1)$-th paragraph) as contextual input, the model proceeds to generate the current paragraph. This use of prior context ensures smooth transitions between paragraphs and contributes to the overall logical flow of the document. \end{itemize}

\subsection{Stage 3: Refine}

The final refinement stage goes beyond superficial edits by systematically evaluating the overall quality of the generated text and identifying specific paragraphs that require targeted revisions.

Specifically, the refinement workflow consists of two key steps:

\begin{itemize}[leftmargin=*,itemsep=0pt,topsep=0pt] \item \textbf{Checker Step:} The model conducts a comprehensive assessment of each paragraph, identifying issues such as logical inconsistencies, unclear expressions, or deviations from the intended narrative structure. \item \textbf{Editor Step:} Based on the feedback from the Checker stage, the model performs precise and targeted modifications to improve textual accuracy, fluency, and structural coherence. \end{itemize}

This iterative and structured refinement process ensures that the final output not only accurately conveys the original intent and narrative objectives but also meets the rigorous standards expected in academic writing.  Appendix~\ref{Stage-3:prompt} provides the detailed prompts for paragraph refine stage.

%% file: 4_2_method.tex
\section{\textit{SuperWriter}-LM}
Following the development of the \textit{SuperWriter}-agent, which introduces structured thinking and iterative discussion mechanisms to significantly enhance text generation quality, we are motivated to explore a central research question:
\emph{Can large language models, when guided by the thinking paradigm provided by the \textit{SuperWriter}-agent data, internalize the ability to generate high-quality long-form content through substantially fewer inference steps—rather than relying on 30 to 40 separate agent calls per sample?} 

To answer this question, we conduct targeted model training experiments. Our objective is not merely to extend output length, but to fundamentally improve coherence, relevance, and depth by distilling the agent’s structured thinking process directly into the model itself. In the following sections, we describe the construction of our high-quality training dataset and the strategic methodology for training LLMs to acquire and internalize the \textit{SuperWriter}-agent’s structured thinking and writing capabilities.


\subsection{SFT-Training}

Our training dataset is sourced from two real-world instruction tuning datasets: English-language and Chinese-language instructions from WildChat-1M~\citep{zhao2024wildchat} and LMSYS-Chat-1M~\citep{zheng2023lmsyschat1m}. To ensure the quality and relevance of the selected instructions for long-form writing tasks, we apply a filtering process using the DeepSeek-R1-Distill-Qwen-32B model~\citep{deepseekai2025deepseekr1incentivizingreasoningcapability} (the filtering method is detailed in Appendix~\ref{app:user_QUERY}). 

\begin{figure}[t]
    \centering
    \includegraphics[width=0.8\linewidth]{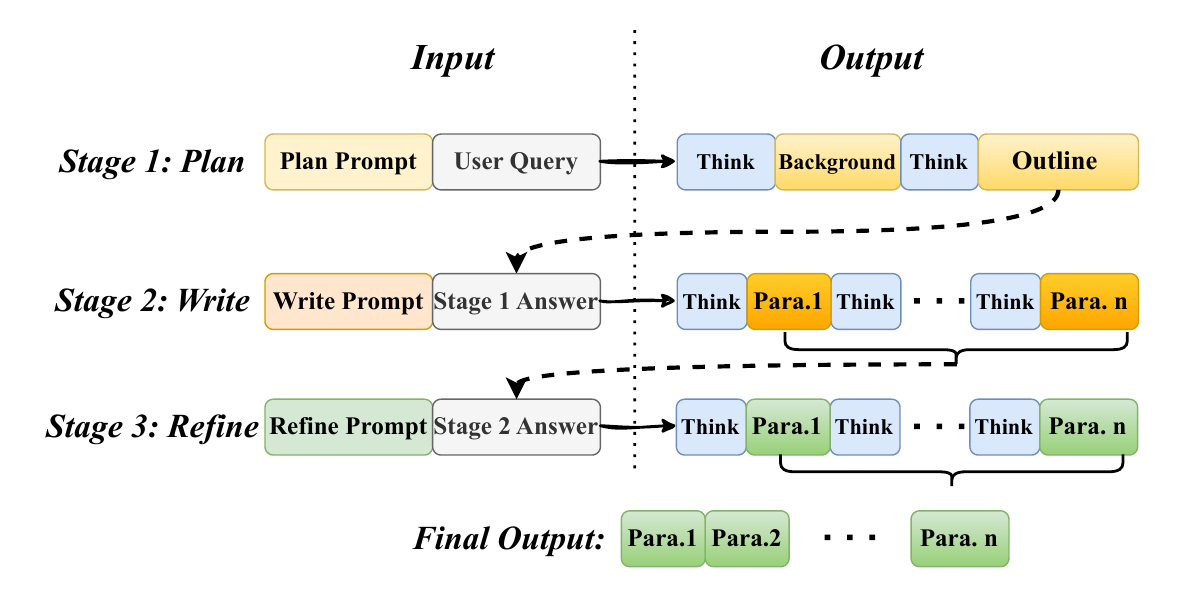}
    \caption{Format of the SFT dataset constructed by the \textit{SuperWriter}-agent. The agent generates training data across three stages—planning, writing, and refining—each incorporating intermediate ``Think'' steps. During inference, the output of Stage 3 is used as the final answer.}
    \label{fig:SFT-data}
\end{figure}

We generate SFT data using the \textit{SuperWriter}-agent (powered by GPT-4o-2024-08-06~\cite{openai2024gpt4ocard}) based on 4,000 filtered instructions.
Each data instance follows a structured pipeline: query $\rightarrow$ outline $\rightarrow$ draft $\rightarrow$ final output. We explicitly segment this pipeline into three stages that align with the internal structure of the \textit{SuperWriter}-agent: \emph{plan} (query $\rightarrow$ outline), \emph{write} (outline $\rightarrow$ draft), and \emph{refine} (draft $\rightarrow$ final output). Rather than training the model directly on full instruction-to-answer SFT pairs, we adopt stage-wise training (illustrated in Fig~\ref{fig:SFT-data}) for two key reasons. First, it better accommodate to real-world user workflows, where users may wish to review and revise intermediate results (e.g., the outline) before progressing to subsequent stages. Second, the complete outputs generated by the agent can be extremely long—some exceeding 100K tokens—posing significant challenges for existing long-context models. By breaking the generation process into stages, we ensure that each training sample remains within 32K tokens, making it more tractable for current models.
Putting them all together, each of the three stages—\emph{plan}, \emph{write}, and \emph{refine}—contains exactly 4,000 data instances, resulting in a total of 12,000 high-quality training data.
During inference, the model performs the generation in three sequential stages to produce the final output.

We train our \textit{SuperWriter}-LM model based on Qwen2.5-7B~\cite{qwen2}, which supports a context window of up to 128K tokens, making it well-suited for long-form text generation. To improve training efficiency, we adopt the packing-based training strategy with loss weighting, as proposed by~\citet{bai2024longalign}.
Training is conducted on a single node equipped with 8×H800 80GB GPUs using DeepSpeed with ZeRO-3 and CPU offloading~\cite{rasley2020deepspeed}. The training setup includes a batch size of 32, a learning rate of $2\times 10^{-5}$, and a context window of 32K tokens. We train the model for four epochs in total.

\subsection{Hierarchical DPO for Multi-Stage Generation}
\label{sec:hier_dpo}

\noindent
  
Direct Preference Optimization~(DPO) has demonstrated effectiveness in aligning policies directly with pairwise human (or proxy-model) preferences for \emph{single-pass} generation tasks~\cite{rafailov2024direct}.  
However, in our scenario, the writing process unfolds sequentially over three distinct stages: \textit{planning}, \textit{drafting}, and \textit{refinement}.  
Applying conventional DPO exclusively to final outputs neglects the valuable preference signals inherently present in earlier stages.  
To address this gap, we introduce a hierarchical multi-stage DPO framework that combines structured preference-data construction with systematic evaluation, inspired by process-annotation methods such as \textsc{TPO}~\cite{liao2025tpo}, \textsc{CHIP}~\cite{fu2025chip}, and Math-Shepherd~\cite{wang-etal-2024-math}.

\begin{figure*}[t]
    \centering
    \includegraphics[width=\linewidth]{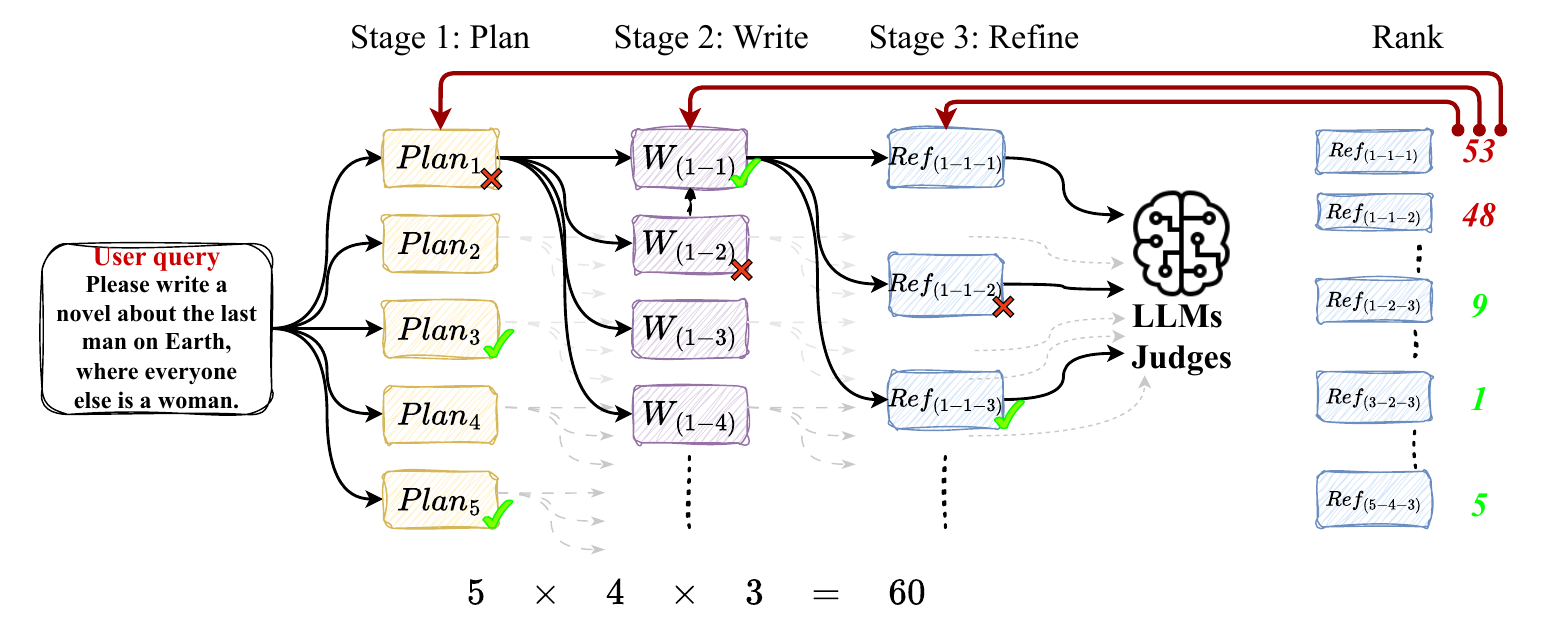}
    \caption{The MCTS begins with 5 distinct writing plans, each leading to 4 written drafts, totaling 20 initial outputs. Each draft is then refined 3 times, resulting in 60 unique final outputs for a single root. A judge LLM ranks these final outputs, and the rankings are back-propagated through the refinement and planning stages. This scoring mechanism helps identify the most effective planning and writing strategies, thereby optimizing the overall writing process.}
    \label{fig:DPO}
\end{figure*}

Specifically, as depicted in Figure~\ref{fig:DPO}, the writing process is structured as a tree explored through Monte-Carlo Tree Search.  
Each path through this tree, labeled $(i,j,k)$, corresponds sequentially to Stage-1 (\textit{plan} $i$), Stage-2 (\textit{draft} $j$), and Stage-3 (\textit{refinement} $k$).  
We embed two key assumptions: well-structured initial \textit{plans} lead to higher-quality draft (Stage-1 Plan $\rightarrow$ Stage-2 Write) and well-refined drafts typically yield better final outputs  (Stage-2 Write $\rightarrow$ Stage-2 Refine). 
Consequently, we \emph{back-propagate} quality signals from leaf nodes (final outputs) upwards through intermediate stages, ensuring the policy learns from decisions at every level rather than only from final outcomes.

\paragraph{Structured evaluation (\textit{Write-judge}).}
To score the final output at each leaf, we introduce \textit{Write-judge}\footnote{Pipeline details and prompts appear in App.~\ref{app:DPO}.}, a six-dimension rubric (0–10 each) chosen from a larger pool of twenty dimensions according to the instruction type (e.g.\ creativity vs.\ logical coherence).  
Following best practice to curb evaluation bias, we use the QwQ-32B model~\cite{qwq32b} to score every output three times under the same temperature settings and take the average.  
Then we propagate scores from the leaf nodes upward to construct DPO pairs as follows.


\paragraph{Step 1: Leaf-score discretization.}
Let $s_{ijk}$ denote the averaged raw evaluation score assigned to leaf $(i,j,k)$, and let $N$ represent the total number of leaf nodes.  
We define the ranking and percentile for each leaf node as follows:
\[
\operatorname{rank}_{ijk} = 1 + \bigl|\{(p,q,r)\mid s_{pqr}>s_{ijk}\}\bigr|,\quad
\pi_{ijk}=\frac{\operatorname{rank}_{ijk}}{N}.
\]

The percentile scores $\pi_{ijk}$ are subsequently discretized into ordinal rewards $r_{ijk}$:
\[
r_{ijk}=
\begin{cases}
+2, & \pi_{ijk}\le\tfrac{1}{6},\\[2pt]
+1, & \tfrac{1}{6}<\pi_{ijk}\le\tfrac{2}{6},\\[2pt]
0,  & \tfrac{2}{6}<\pi_{ijk}\le\tfrac{4}{6},\\[2pt]
-1, & \tfrac{4}{6}<\pi_{ijk}\le\tfrac{5}{6},\\[2pt]
-2, & \pi_{ijk}>\tfrac{5}{6}.
\end{cases}
\]

\paragraph{Step 2: Reward aggregation.}
Ordinal rewards are propagated upwards through the tree by averaging across child nodes at each intermediate stage:
\begin{alignat*}{2}
\hat r_{ij}      &=\frac{1}{|\{k\}|}\sum_{k} r_{ijk}       &\quad &\text{(Stage 2 node)}\\[3pt]
\hat r_{i}       &=\frac{1}{|\{j\}|}\sum_{j} \hat r_{ij}   &      &\text{(Stage 1 node)}
\end{alignat*}

\paragraph{Step 3: Preference-pair construction.}
We harvest pairwise preferences at \emph{every} hierarchy level, ensuring that each decision contributes to the optimization at every step.

\begin{enumerate}[leftmargin=1.9em,itemsep=4pt,topsep=2pt,label=\textbf{L\arabic*.}]
\item \textbf{Stage 1 (Plan).}\;Let $B=\arg\max_i \hat r_i$ be the best plan and $W_1,W_2$ the worst two.  
      \[
        P_1=\{(B,W_1),(B,W_2)\}
      \]

\item \textbf{Stage 2 (Write).}\;For the best two plans $i\in\{B,S\}$ (with $S$ being the second best),
      \[
        P_2(i)=\bigl(\arg\max_j\hat r_{ij},\,\arg\min_j\hat r_{ij}\bigr),\qquad
        P_2=\bigcup_i P_2(i).
      \]

\item \textbf{Stage 3 (Refine).}\;For each $(i,j^{\star})\in P_2$,
      \[
        P_3(i,j^{\star})=\bigl(\arg\max_k s_{ijk},\,\arg\min_k s_{ijk}\bigr),\qquad
        P_3=\bigcup_{(i,j^{\star})}P_3(i,j^{\star}).
      \]
\end{enumerate}

The complete preference dataset is therefore
\[
  \mathcal{D}_{\text{DPO}} \;=\; P_1 \cup P_2 \cup P_3.
\]

\paragraph{Optimization objective.}
Finally, we optimize the policy~$\pi_\theta$ using the standard DPO loss:
\[
\mathcal{L}_{\text{DPO}}
\;=\;
-\,\mathbb{E}_{(x,y^{+},y^{-})\sim\mathcal{D}_{\text{DPO}}}
  \Bigl[
    \log \sigma\!\bigl(\,\beta\,[\,s_{\theta}(x,y^{+}) - s_{\theta}(x,y^{-})]\bigr)
  \Bigr].
\]

Using the above method, we obtained a DPO preference dataset and employed 360-LLaMA-factory~\cite{360-llama-factory} to continue context-parallel DPO training of the supervised fine-tuned \textit{SuperWriter}-LM with a batch size of 32 and a learning rate of $1\times 10^{-6}$.

%% file: 5_experiment.tex
\section{Experiment}

\subsection{Experiment Setup}

\paragraph{Inference Setup.} We utilize the SGLang~\citep{zheng2023sglang} system, which optimizes key-value (KV) cache memory for efficient large-scale inference. This system is crucial for handling long-form generation efficiently, reducing memory overhead, and maximizing inference throughput. Inferences are performed using BFloat16 precision on $8 \times$NVIDIA H800 GPUs. This setup ensured consistency and efficiency in the inference process. 

\input{tabel/main_tab}

\paragraph{Benchmark Setup.}

We conduct our experiments on two datasets. The first is WritingBench~\cite{wu2025writingbenchcomprehensivebenchmarkgenerative}, a comprehensive benchmark designed to evaluate large language models across six major writing domains and 100 sub-domains, encompassing creative, persuasive, informational, and technical tasks. The benchmark comprises 1,200 real-world writing prompts, each paired with five instance-specific evaluation criteria. WritingBench adopts a query-dependent evaluation framework and leverages a Qwen2.5-7B critic model fine-tuned on 50K human-labeled samples to produce fine-grained assessments of style, format, and length, achieving an 83\% agreement rate with human judgments.

The second dataset includes approximately 200 real-world user instructions manually curated by us. We evaluate model performance through pairwise comparisons across several baselines (Win-rate), including LongWriter-8B~\cite{bai2024longwriter}, DeepSeek-R1-Distill-Qwen-7B~\cite{deepseekai2025deepseekr1incentivizingreasoningcapability}, Writing-Model-Qwen-7B~\cite{wu2025writingbenchcomprehensivebenchmarkgenerative}\footnote{This refers to the SFT model obtained via rejection sampling using the WritingBench critic model.}, Qwen3~\cite{qwen2.5}, LLaMA-4-Scout~\cite{meta2025llama4}, DeepSeek-V3~\cite{deepseekai2025deepseekv3technicalreport}, and DeepSeek-R1~\cite{deepseekai2025deepseekr1incentivizingreasoningcapability}. All models are evaluated using consistent decoding configurations (temperature = 0.6, top-p = 0.95), and win rates are calculated to quantify comparative performance.

\subsection{WritingBench Result}

We evaluate our 7B sized \textit{SuperWriter}-LM model on the WritingBench benchmark. As shown in Tab~\ref{tab:main table}, \textit{SuperWriter}-LM achieves an overall performance (Avg) of 8.51—second-best among all models, closely following DeepSeek-R1~\cite{deepseekai2025deepseekr1incentivizingreasoningcapability}. It demonstrates strong capabilities in both Chinese--8.6 and English--8.5, even matching DeepSeek-R1 in English performance.

Across different domains, \textit{SuperWriter}-LM achieves the highest scores in three major areas: (D1) Academic \& Engineering--8.6, (D2) Finance \& Business--8.7, (D3) Politics \& Law--8.7 and (D5) Education--8.7, even slightly outperforming the DeepSeek-R1  model. Additionally, \textit{SuperWriter}-LM satisfies various special writing requirements, except the length\_C setting. This discrepancy is primarily due to the nature of agent-generated data, which tends to produce longer outputs even for short-text tasks—an issue that does not affect long-form generations.

\subsection{Win-Rate Result}
\begin{figure*}[t]
    \centering
    \includegraphics[width=1\linewidth]{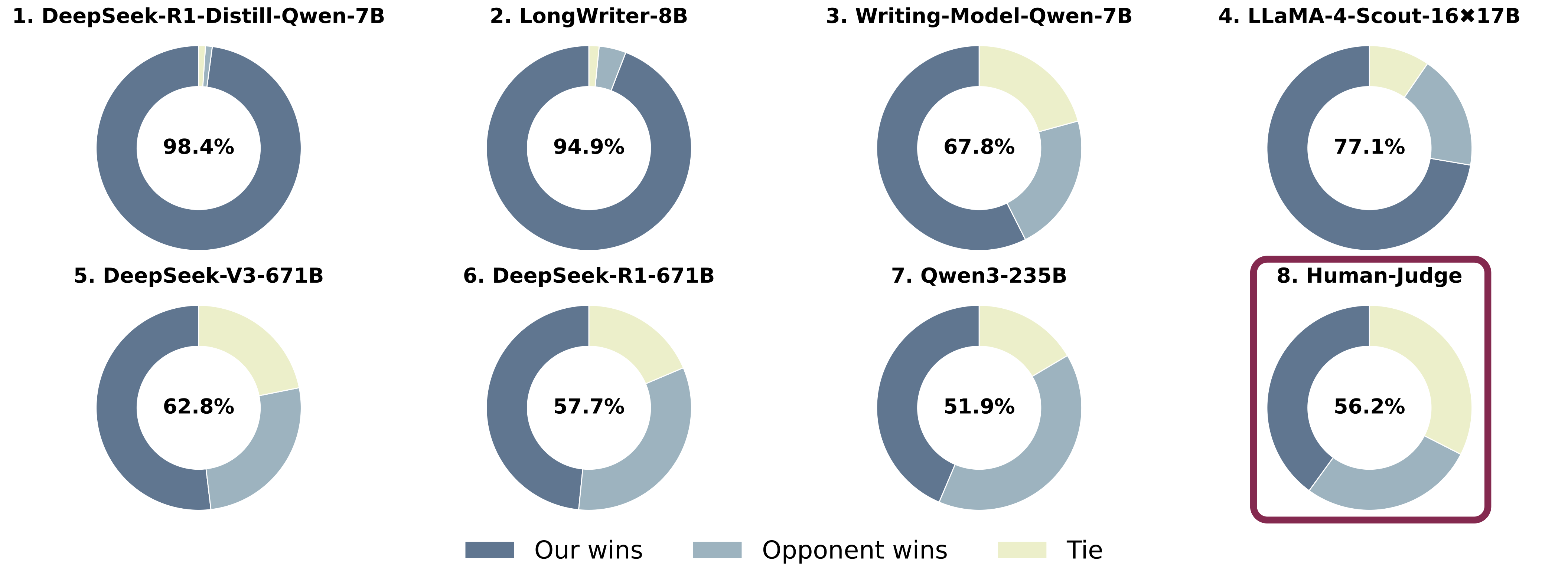}
    \caption{This figure presents eight donut charts comparing the win rates of our model against seven different baselines. The win-rate in the 8th chart is evaluated by human (Ours vs. Writing-Model-Qwen-7B) while the rest are evaluated by GPT-4.1. Each chart shows the proportions of wins, losses, and ties, with the central percentage indicating the overall win rate. When calculating the win rate, a tie is counted as 0.5 win for both sides.}
    \label{fig:winrate}
\end{figure*}

WritingBench adopts a critic model to evaluate model outputs by assigning scores (ranging from 1 to 10) across 4–5 distinct dimensions. However, this evaluation approach has several limitations. First, due to the relatively small size of the critic model, it may be vulnerable to some word/sentence hacking. Second, we observe that WritingBench primarily focuses on formal or professional writing tasks—such as summaries and reports—whereas our analysis of real-world data from \textit{WildChat}~\cite{zhao2024wildchat} and \textit{LMSYS-Chat-1M}~\cite{zheng2023lmsyschat1m} reveals that creative writing (e.g., storytelling, fiction) constitutes a significant portion of user queries.
To address these concerns, we adopt a more direct and interpretable evaluation metric: \textbf{win-rate}. We evaluate model performance on nearly 200 real-world user queries collected from the aforementioned datasets. For each query, responses are generated by \textit{SuperWriter}-LM and six baseline models.

\paragraph{LLM Evaluation.} For automatic evaluation, we adopt LLM-as-a-judge~\cite{bai2023benchmarking,zheng2023judging} to perform pairwise comparison using \texttt{GPT-4.1-2025-04-14}~\cite{openai2025gpt41}.
To mitigate positional bias, we conduct two evaluations per pair by swapping the response order: \texttt{\texttt{Evaluation\_Prompt + A + B}} and \texttt{\texttt{Evaluation\_Prompt + B + A}}. Based on the two judgments, we categorize the results into win, loss, or tie and compute win-rate results\footnote{Evaluation prompts are provided in Appendix \ref{app:winrate_prompt}.}.

As shown in Figure~\ref{fig:winrate}, \textit{SuperWriter}-LM demonstrates a substantial performance lead among models of the same size (models 1, 2, and 3). Furthermore, in comparisons against larger-sized models (models 4, 5, 6, and 7), \textit{SuperWriter}-LM remains competitive and, in some cases, even slightly outperforms state-of-the-art LLMs. Taken together, these results suggest that \textit{SuperWriter}-LM sets a new performance benchmark among 7B-scale models and even challenges the capabilities of existing state-of-the-art systems. We also present several case studies in Appendix \ref{app:case_study}.

\paragraph{Human Evaluation.}

To mitigate potential inaccuracies in automatic evaluation, we conduct an human supplementary assessment on approximately 200 real-world user queries, comparing \textit{SuperWriter}-LM with Writing-Model-Qwen-7B~\cite{wu2025writingbenchcomprehensivebenchmarkgenerative}. For each query, three independent annotators with undergraduate degrees. were tasked with evaluating and determining the preferred response, with outcomes categorized as win, loss, or tie—following the same standard as the automatic evaluation. The aggregated results are shown in Figure~\ref{fig:winrate} (8), and \textit{SuperWriter}-LM demonstrates stronger performance under human judgment. However, due to the annotators’ tendency to assign a tie when the differences between two responses are subtle, the overall win rate appears slightly lower.

\subsection{Ablation Study}

\begin{wrapfigure}[10]{r}{0.5\textwidth}
  \centering
  \small
  \begin{tabular}{l c c c}
    \toprule
    Model                   & Avg  & ZH  & EN  \\
    \midrule
    Qwen2.5-7B-Instruct       & 7.43 & 7.3 & 7.5 \\
    + \textit{SuperWriter final output}                & 8.21 & 8.3 & 8.2 \\
     + \textit{Three-Stage}             & \underline{8.47} & \underline{8.5} & \underline{8.4} \\
     + \textit{Hierarchical DPO}   & \textbf{8.51} & \textbf{8.6} & \textbf{8.5} \\
    \bottomrule
  \end{tabular}
  \caption{Performance comparison on WritingBench~\cite{wu2025writingbenchcomprehensivebenchmarkgenerative} -- Avg, ZH and EN.}
  \label{tab:Ablation Study}
\end{wrapfigure}
Finally, we conduct an ablation study comprising four different setups evaluated on the WritingBench benchmark. The first setup uses the base model, \textit{Qwen2.5-Instruct}, as the performance baseline. The second setup, \textit{SuperWriter-final-answer}, takes the user query as input and produces the final output from the Stage-3 \textit{refine} step of the \textit{SuperWriter}-agent---this is a one-pass generation without any explicit thinking process and achieves an average score of 8.21 (ZH: 8.3, EN: 8.2). The third setup, \textit{+Three-Stage}, corresponds to our SFT-trained model, which explicitly performs planning, drafting, and refining in a chained, multi-stage manner, incorporating structured thinking. This setup further improves performance to 8.47 (ZH: 8.5, EN: 8.4). The final setup is the full model further enhanced with our hierarchical DPO optimization, reaching the highest score of 8.51 (ZH: 8.6, EN: 8.5). As shown in Table~\ref{tab:Ablation Study}, each additional component leads to consistent performance improvements, demonstrating the effectiveness of our proposed approach in structured writing tasks.

%% file: tabel/main_tab.tex
\begin{table*}[t]
\centering 
\footnotesize 
\resizebox{1.0\textwidth}{!}{
\begin{tabular}{l|c|cc|cccccc|clclcl} 
\toprule 
\multirow{2.5}{*}{\textbf{Models}} & \multirow{2.5}{*}{\textbf{Avg}} & \multicolumn{2}{c|}{\textbf{Languages}} & \multicolumn{6}{c|}{\textbf{Domains}} & \multicolumn{6}{c}{\textbf{Requirements}} \\ 
\cmidrule(l){3-16} 
& & \textbf{ZH} & \textbf{EN} & \textbf{D1} & \textbf{D2} & \textbf{D3} & \textbf{D4} & \textbf{D5} & \textbf{D6} & \textbf{R1} & \multicolumn{1}{c|}{\textbf{C}} & \textbf{R2} & \multicolumn{1}{c|}{\textbf{C}} & \textbf{R3} & \multicolumn{1}{c}{\textbf{C}} \\ 

\midrule 
\multicolumn{16}{l}{\cellcolor{lightblue}\textit{Proprietary LLMs}} \\ 
\midrule 
ChatGPT-4o-latest & 8.16 & 8.3 & 8.1 & 8.1 & 8.1 & 8.2 & 8.1 & 8.4 & 8.1 & 8.3 & 8.7 & 8.2 & 8.9 & 8.2 & 8.3 \\
o1-Preview & 8.15 & 8.1 & 8.2 & 8.0 & 8.1 & 8.2 & 8.2 & 8.4 & 8.1 & 8.2 & 8.6 & 8.2 & 8.8 & 8.2 & 8.2 \\
Claude-3.5-Sonnet & 7.71 & 7.7 & 7.7 & 7.6 & 7.5 & 7.6 & 7.7 & 7.9 & 8.0 & 7.9 & 8.5 & 7.7 & 8.5 & 7.9 & 8.0 \\
Gemini-1.5-Pro & 7.78 & 7.8 & 7.7 & 7.7 & 7.5 & 7.8 & 7.9 & 8.0 & 7.9 & 7.9 & 8.6 & 7.9 & 8.8 & 7.9 & 8.0 \\
Qwen-Max & 8.37 & 8.4 & 8.3 & 8.3 & 8.3 & 8.4 & \underline{8.4} & \underline{8.5} & 8.4 & 8.5 & 8.7 & 8.4 & \textbf{9.0} & 8.4 & 8.5 \\
\midrule 
\multicolumn{16}{l}{\cellcolor{mediumblue}\textit{Open-source LLMs}} \\
\midrule 
Deepseek-R1                & \textbf{8.55} & \textbf{8.7} & \textbf{8.5} & \underline{8.5} & \underline{8.5} & \underline{8.6} & \textbf{8.6} & \textbf{8.7} & \textbf{8.6} & \textbf{8.7} & \textbf{8.9} & \textbf{8.6} & \textbf{9.0} & \textbf{8.6} & \textbf{8.7} \\
Deepseek-V3 & 7.95 & 8.0 & 7.9 & 7.9 & 7.8 & 8.0 & 7.8 & 8.2 & 8.0 & 8.1 & 8.6 & 8.0 & 8.9 & 8.0 & 8.2 \\
Mistral-Large-Instruct & 7.64 & 7.6 & 7.7 & 7.7 & 7.6 & 7.8 & 7.3 & 7.9 & 7.6 & 7.7 & 8.2 & 7.7 & 8.7 & 7.7 & 7.9 \\
Qwen-2.5-72B-Instruct & 7.90 & 8.0 & 7.9 & 8.0 & 7.8 & 8.1 & 7.7 & 8.2 & 7.8 & 8.0 & 8.3 & 8.0 & 8.8 & 7.9 & 8.0 \\
Qwen-2.5-7B-Instruct & 7.43 & 7.3 & 7.5 & 7.7 & 7.4 & 7.6 & 6.9 & 7.8 & 7.3 & 7.5 & 7.9 & 7.6 & 8.6 & 7.4 & 7.5 \\
Llama-3.3-70B-Instruct & 7.01 & 6.7 & 7.3 & 7.0 & 6.9 & 7.0 & 6.8 & 7.3 & 7.3 & 7.1 & 7.8 & 7.1 & 8.2 & 7.0 & 7.2 \\
Llama-3.1-8B-Instruct & 6.35 & 5.7 & 6.9 & 6.6 & 6.4 & 6.1 & 6.0 & 6.7 & 6.6 & 6.4 & 7.0 & 6.4 & 7.6 & 6.3 & 6.4 \\
\midrule 
\multicolumn{16}{l}{\cellcolor{deepblue}\textit{Capability-enhanced LLMs}} \\
\midrule 
Suri-I-ORPO & 4.97 & 4.4 & 5.5 & 5.6 & 5.3 & 5.0 & 4.1 & 5.0 & 5.1 & 4.8 & 5.2 & 5.0 & 5.4 & 4.5 & 4.0 \\
LongWriter-8B & 7.91 & 7.9 & 7.9 & 8.0 & 8.1 & 8.1 & 7.7 & 8.1 & 7.6 & 7.9 & 8.2 & 8.1 & 8.8 & 7.7 & 7.7 \\
Writing-Model-Qwen-7B & 8.49 & \underline{8.6} & 8.4 & 8.4 & 8.4 & 8.6 & 8.4 & 8.6 & \underline{8.5} & 8.6 & 8.8 & 8.5 & 9.0 & 8.5 & 8.6 \\
Writing-Model-Llama-7B & 8.49 & \underline{8.6} & 8.4 & 8.5 & 8.4 & 8.6 & 8.4 & 8.6 & \underline{8.5} & 8.6 & 8.8 & 8.5 & 8.9 & 8.5 & 8.5 \\
\textbf{\emph{SuperWriter}-LM (ours)}& \underline{8.51} & \underline{8.6} & \textbf{8.5} & \textbf{8.6} & \textbf{8.7} & \textbf{8.7} & 8.2          & \textbf{8.7} & 8.2          & 8.4          & 8.5          & \textbf{8.6} & 8.4          & 8.0          & 6.3          \\

\bottomrule 
\end{tabular}}
\caption{WritingBench performance of different LLMs across 6 domains and 3 writing requirements evaluated with our critic model (scale: 1-10). The six domains include: (D1) Academic \& Engineering, (D2) Finance \& Business, (D3) Politics \& Law, (D4) Literature \& Art, (D5) Education, and (D6) Advertising \& Marketing. The three writing requirements assessed are: (R1) Style, (R2) Format, and (R3) Length. Here, ``C'' indicates category-specific score. }
\label{tab:main table}
\end{table*}

%% file: 3_related_work.tex
\section{Related Work}

\paragraph{Long-context Language Models}
Recent research on long-context language models (LLMs) has primarily focused on extending the input context length, enabling models to process substantially longer inputs. Approaches in this direction generally fall into two categories. The first involves zero-shot techniques, aiming to expand the model's receptive field without additional training~\citep{han2023lm,xiao2023efficient,zhang2024soaring,jin2024llm,an2024training}. The second category encompasses fine-tuning-based methods, where models undergo training on extended sequences to explicitly enhance their ability to handle long contexts~\citep{chen2023extending,peng2023yarn,xiong2024effective,chen2023longlora,bai2024longalign,fu2024data}. However, existing studies have predominantly concentrated on input expansion, often overlooking the critical necessity for corresponding output capabilities. Recent empirical findings by \citet{bai2024longwriter} and~\citet{quan2024language} have shown a significant mismatch between maximum input lengths (over 100K tokens) and achievable output lengths (approximately 2K words). 

\paragraph{Long-form Text Generation}
Recent advancements emphasize architectural innovations and specialized training paradigms~\citep{salemi2025experteffectiveexplainableevaluation,que2024hellobenchevaluatinglongtext,liu-etal-2023-task,Li2023TeachLT}. For instance, Re3 employs a recursive prompting strategy, effectively maintaining narrative coherence in extended storytelling tasks~\citep{yang-etal-2022-re3}. Other approaches, such as DOC~\citep{yang-etal-2023-doc} and hierarchical outlining methods~\citep{wang2024generatinglongformstoryusing}, leverage structured task decomposition to improve overall narrative structure and content coherence. Recently, personalization has emerged as another crucial dimension, with models like LongLaMP~\citep{kumar2024longlampbenchmarkpersonalizedlongform} and reasoning-enhanced self-training approaches~\citep{salemi2025reasoningenhancedselftraininglongformpersonalized} adapting outputs according to individual user preferences. Additionally, long-form question-answering techniques have gained attention for addressing complex and detailed queries with comprehensive responses~\citep{dasigi-etal-2021-dataset,stelmakh-etal-2022-asqa,pmlr-v202-lee23n,tan-etal-2024-proxyqa}. Despite these advancements, existing methods often suffer from limitations such as inconsistent coherence, lack of theoretical grounding, and challenges in generating consistently high-quality content at large scales~\citep{wu2024longgenbench,que2024hellobenchevaluatinglongtext}. In response, our work addresses these limitations by proposing a theoretical agent framework for long-form writing, accompanied by novel training methodologies designed specifically to enhance the model’s capability in generating coherent and extended texts.

%% file: 6_conclusions.tex
\section{Conclusion}

\textit{SuperWriter} addresses the challenge of long-form text generation by introducing a structured writing process—planning, writing, and refining—guided by the \textit{SuperWriter}-agent. This approach teaches the model to ``think before writing'' and produces high-quality supervision signals. Combined with a hierarchical DPO strategy, the model learns to align its output across all writing stages.

Experiments show strong results: \textit{SuperWriter}-LM outperforms all same-size models on WritingBench and even exceeds the 671B DeepSeek-R1 model~\cite{deepseekai2025deepseekr1incentivizingreasoningcapability} in key domains. It also wins over 98\% of real-user comparisons against top open-source baselines. These results confirm the value of multi-stage generation and structured preference learning for improving writing quality.

\section{Limitation}

While \textit{SuperWriter}-LM demonstrates strong performance in long-form text generation, several limitations remain:

\textbf{(1) Inference latency.}
Compared to single-pass generation models such as \textit{LongWriter}~\cite{bai2024longwriter} or \textit{Suri}~\cite{pham2024suri}, our method incurs additional inference time due to its three-stage Framework. Although this is significantly more efficient than multi-round agent-based pipelines (e.g., requiring 30–40 calls per output), the structured \textit{plan → write → refine} process still requires three sequential forward passes, which may increase user-perceived latency in real-world applications. 

\textbf{(2) Model scale.}  
Our current implementation is built on a 7B parameter backbone (Qwen2.5)\cite{qwen2025qwen25technicalreport}, which strikes a balance between performance and cost. However, this moderate scale may limit the model’s internal world knowledge, particularly in knowledge-intensive or specialized writing scenarios (e.g., legal, medical, and scientific domains). In our qualitative analysis, some outputs exhibited shallow factual grounding or subtle reasoning errors. 

\textbf{(3) Lack of online reinforcement learning.} 
Lack of online reinforcement learning.
Our alignment stage relies solely on offline Direct Preference Optimization (DPO)~\cite{rafailov2024direct}, trained from static preference pairs. While effective, this setup lacks the adaptivity of online Reinforcement Learning from Human Feedback (RLHF)~\cite{christiano2023deepreinforcementlearninghuman}, which allows models to continually refine outputs through exploration. The key bottleneck is the high rollout cost when applying general-purpose reward models to long outputs. Designing scalable, low-latency reward models or reward distillation methods~~\cite{bai2022training,ouyang2022training} for long-form tasks is thus a promising direction for future research.

%% file: 7_APPENDIX.tex
\newpage
\section{Appendix}
\input{8_Agent_prompt_appendix}

\subsection{Evaluation Prompt for  Hierarchical DPO}
\label{app:DPO}

To support structured preference data construction for Direct Preference Optimization (DPO), we design and apply a sequence of modular prompts. Each serves a specific role within the evaluation pipeline. Below is an overview of their usage:

\label{sec:eval-prompts-logic}

The evaluation pipeline comprises the following steps:

\begin{enumerate}[leftmargin=*,itemsep=0pt,topsep=0pt]
    \item \textbf{Rubric Definition} (\texttt{evaluation\_criteria}): Defines the complete set of General and Special evaluation dimensions. This rubric is reused across all queries.

    \item \textbf{Criterion Selection Schema} (\texttt{format\_query}): Specifies the JSON format for selecting six criteria (three General, three query-relevant Special) and rewriting their Definitions and Standards to match the specific query context.

    \item \textbf{Criterion Selection Prompt}: Combines the rubric and schema to instruct the model to select and customize criteria. The output is a JSON object referred to as \texttt{evaluate\_standard}.

    \item \textbf{Scoring Format Schema} (\texttt{format\_eval}): Specifies the expected evaluation output format: for each selected criterion, the model must return an \textit{Analysis} string and a numeric \textit{Score}.

    \item \textbf{Final Scoring Prompt}: Provides the model with a query, its generated result, the customized \texttt{evaluate\_standard}, and the \texttt{format\_eval} schema. The model performs criterion-wise evaluation and outputs a structured JSON.
\end{enumerate}

\noindent\textbf{Outcome:} This pipeline yields structured, query-specific evaluations that are interpretable, machine-parsable, and suitable for training with DPO loss.

\begin{tcolorbox}[colback=gray!5!white, colframe=black!75!black, title=\texttt{evaluation\_criteria Prompt}, fonttitle=\bfseries, breakable]
\footnotesize
Evaluation Criteria

\textbf{1. General Criteria (Applicable to All Genres)}

\textbf{1.1 Relevance} \\
\textbf{Definition}: How well the content matches the user’s request, and whether it addresses the intended purpose or topic. \\
\textbf{Standards}: \\
\quad \textbf{10}: Fully aligned with the user’s needs, highly relevant to the request. \\
\quad \textbf{7--9}: Mostly relevant, with some minor deviations or less-than-perfect alignment. \\
\quad \textbf{4--6}: Partially relevant, with the majority of the content not matching the user’s request. \\
\quad \textbf{1--3}: Completely irrelevant, fails to meet the user’s needs.

\vspace{0.5em}
\textbf{1.2 Coherence} \\
\textbf{Definition}: The clarity of the structure, and the logical flow of ideas and transitions between paragraphs and sentences. \\
\textbf{Standards}: \\
\quad \textbf{10}: Clear structure, strong logical progression, and smooth flow of content with natural transitions. \\
\quad \textbf{7--9}: Mostly coherent, with minor lapses or jumps in logic. \\
\quad \textbf{4--6}: Structure is somewhat unclear, and there are noticeable gaps or jumps in logic. \\
\quad \textbf{1--3}: Disorganized and confusing, with no clear structure or logical connections.

\vspace{0.5em}
\textbf{1.3 Clarity} \\
\textbf{Definition}: How clear and easy the content is to understand, and whether it includes sufficient detail. \\
\textbf{Standards}: \\
\quad \textbf{10}: Clear expression, rich in detail, and easy to understand. \\
\quad \textbf{7--9}: Mostly clear, but may have some lengthy or unclear parts. \\
\quad \textbf{4--6}: Expression is somewhat muddled, lacks detail, and is difficult to understand. \\
\quad \textbf{1--3}: Extremely vague, minimal information, and difficult to comprehend.

\vspace{0.5em}
\textbf{2. Special Criteria (Applicable to Specific Genres)}

\textbf{2.1 Creativity and Uniqueness} \\
\textbf{Definition}: Whether the content is innovative, offering new perspectives, or showcasing original expression. \\
\textbf{Standards}: \\
\quad \textbf{10}: Highly creative and unique, presenting entirely new or unconventional ideas or perspectives. \\
\quad \textbf{7--9}: Creative, but some parts are more conventional or lack originality. \\
\quad \textbf{4--6}: Limited creativity, mostly traditional content with little innovation. \\
\quad \textbf{1--3}: Lacks creativity, offering conventional or uninspired content.

.....

\end{tcolorbox}

\begin{tcolorbox}[colback=gray!5!white, colframe=black!75!black, title=\texttt{format\_eval Prompt}, fonttitle=\bfseries, breakable]
\footnotesize
The final output should be in JSON format, structured as follows: \verb|```json| \\
\{\\
\quad "Criterion 1": \{\\
\quad\quad "Analysis": "Analysis content",\\
\quad\quad "Score": X\\
\quad \},\\
\quad "Criterion 2": \{\\
\quad\quad "Analysis": "Analysis content",\\
\quad\quad "Score": X\\
\quad \},\\
\quad ...\\
\}\\
\verb|```|
\end{tcolorbox}

\begin{tcolorbox}[colback=gray!5!white, colframe=black!75!black, title=\texttt{format\_query Prompt}, fonttitle=\bfseries, breakable]
\footnotesize
After think give final answer in JSON format. Structured as follows: \verb|```json| \\
\{\\
\quad "Selected Criterion 1": \{\\
\quad\quad "Definition": "Criterion definition",\\
\quad\quad "Standards": "Scoring standards for the query, give a simple rubric from 1 to 10, 10: ..., 7-9: ..., 4-6: ..., 1-3: ..."\\
\quad \},\\
\quad "Selected Criterion 2": \{\\
\quad\quad "Definition": "Criterion definition",\\
\quad\quad "Standards": "..."\\
\quad \},\\
\quad ...\\
\}\\
\verb|```|
\end{tcolorbox}

\begin{tcolorbox}[colback=gray!5!white, colframe=black!75!black, title=\texttt{prompt}, fonttitle=\bfseries, breakable]
\footnotesize
Please refer to the evaluation criteria outlined below:

\texttt{\{evaluation\_criteria\}}

\textbf{Task:}

You are tasked with evaluating the query: \verb|`{query}`|. From the \texttt{"Special Criteria"} section, select 3 relevant criteria, and from the \texttt{"General Criteria"} section, select all criteria (Relevance, Coherence, Clarity), for a total of 6 criteria.

Be sure to: \\
1. Think step-by-step about why each criterion is relevant to the query. \\
2. Think step-by-step through the query and how each criterion applies. \\
3. Provide a brief analysis for each selected criterion on how it applies to the query. \\
4. Integrate the above reasoning into the \texttt{Definition} and \texttt{Standards} sections of each criterion.

\texttt{\{format\_query\}}
\end{tcolorbox}
\begin{tcolorbox}[colback=gray!5!white, colframe=black!75!black, title=\texttt{prompt}, fonttitle=\bfseries, breakable]
\footnotesize
\#\#\# Query: \{query\}

\#\#\# Result: \texttt{<start>} \{clean\_res\} \texttt{<end>}

\#\#\# Evaluation Standard: \texttt{\{json.dumps(evaluate\_standard, ensure\_ascii=False)\}}

Based on the provided info, perform a rigorous evaluation. \{format\_eval\}
\end{tcolorbox}

\subsection{Evaluation Prompt for Win-Rate Judgment}
\label{app:winrate_prompt}
\begin{tcolorbox}[colback=gray!5!white, colframe=black!75!black, title=\textbf{SYSTEM\_PROMPT for GPT-4.1 Win-Rate Evaluation}, breakable]
\small
Please act as an impartial judge and evaluate the quality of the written responses provided by two AI assistants to the user’s writing prompt below. You will be given Assistant A’s response and Assistant B’s response. Your job is to determine which assistant's writing is superior.

Evaluate them on the following criteria: \\
\textbf{1. Relevance and Completeness}: Does the assistant fully respond to the writing prompt? Does the length meet the user's query expectations? Is the content relevant to the topic, and does it provide sufficient depth, length, and detail, rather than drifting off-topic or simplistic? \\
\textbf{2. Writing Quality}: Evaluate whether the assistant's writing is clear, fluent, and free of obvious grammatical errors. The overall quality of the writing is high, with elegant. \\
\textbf{3. Creativity and Originality}: If applicable, assess the creativity of the response. Does the assistant offer fresh perspectives, unique insights, or demonstrate a certain level of originality? \\
\textbf{4. Specificity and Detail}: Determine whether the assistant provides concrete examples or detailed explanations. Properly justified repetition is permissible. \\
\textbf{5. Tone and Style}: Is the tone appropriate for the writing prompt? Is the writing style consistent throughout? Consider whether it aligns with the expectations of the intended audience or writing purpose.

After evaluating each response, determine which one is superior based on the factors above. Provide your explanation and then select one of the following final verdicts:

\begin{itemize}
    \item Assistant A is significantly better: \texttt{[[A>>B]]}
    \item Assistant A is slightly better: \texttt{[[A>B]]}
    \item Tie, relatively the same: \texttt{[[A=B]]}
    \item Assistant B is slightly better: \texttt{[[B>A]]}
    \item Assistant B is significantly better: \texttt{[[B>>A]]}
\end{itemize}

Example output: \texttt{My final verdict is tie: [[A=B]]}.
\end{tcolorbox}

\subsection{Case Study}
\label{app:case_study}
\begin{tcolorbox}[
  colback=blue!5!white,      
  colframe=blue!30!white,    
  boxrule=0.5pt,
  title=Research on the Impact of Digital Inclusive Finance on Household Financial Asset Allocation,
  fonttitle=\bfseries,
  coltitle=black,
  sharp corners=south,
  arc=2mm,
  top=2mm, bottom=2mm, left=2mm, right=2mm, breakable
]
\textbf{Query:} Research on the Impact of Digital Inclusive Finance on Household Financial Asset Allocation.

\bigskip

\textbf{\emph{SuperWriter}-LM Response:} Digital inclusive finance (DIF) has become a cornerstone of global financial innovation, transforming access to financial services for millions. Over 70\% of Sub-Saharan Africa now uses mobile money platforms like Kenya’s M-Pesa, while India’s Digital India initiative has spurred a 40\% rise in crypto adoption since 2020. These advancements highlight DIF’s transformative potential, yet existing research has fallen short in analyzing three critical dimensions: the mechanisms driving shifts in household asset allocation, the equity trade-offs between access and exclusion, and the risk mitigation required to balance innovation with stability. Most studies focus on access expansion or behavioral finance theories in isolation, neglecting how DIF’s design—such as gamified investing apps or frictionless cross-border transfers—affects portfolio composition, risk tolerance, or wealth gaps. Methodological gaps persist too: cross-sectional data often conflate correlation with causation, and equity analyses rarely account for cultural or infrastructural moderators like trust in institutions or gender norms. This study addresses these gaps by rigorously examining how DIF influences household financial asset allocation, including diversification, risk-return preferences, and equity access. Grounded in behavioral finance (e.g., FUDT), financial inclusion theory, and institutional economics, it employs a mixed-methods approach combining quantitative surveys (n=10,000 households in Nigeria and Indonesia), qualitative interviews with users and regulators, and ethnographic observations. Grounded in these theories, the study’s five hypotheses guide the analysis: (1) DIF adoption increases formal market participation, (2) behavioral nudges shift portfolios toward riskier assets, (3) DIF reduces wealth gaps for underserved groups, (4) geographic/cultural factors moderate these effects, and (5) DIF amplifies exposure to volatile/high-risk products. By bridging interdisciplinary insights—from behavioral biases to institutional trust—the study offers actionable strategies for policymakers and fintech firms to design equitable, risk-aware systems. Its findings aim to inform global debates on how to harness DIF’s potential while mitigating its pitfalls, ensuring financial innovation serves as a ladder out of poverty rather than a trap into risk.

Empirical studies have highlighted DIF’s role in expanding financial access. For instance, Kenya’s M-Pesa reduced informal savings reliance by 20\% among rural households (Ratha, 2012), while India’s Digital India initiative spurred a 40\% rise in crypto adoption since 2020 (Reserve Bank of India, 2023). These findings underscore DIF’s potential to bridge wealth gaps and democratize financial inclusion. However, such studies focus predominantly on access metrics—such as account ownership or transaction frequency—rather than how this access translates to \textbf{changes in household financial behavior}, such as asset allocation or risk tolerance. By including unregulated peer-to-peer platforms in DIF metrics, prior studies may overstate the service’s equity impact, as informal tools often lack safeguards against predatory practices or volatility. This conflation also masks how DIF’s design—such as gamified apps or cross-border transfers—affects risk tolerance or wealth gaps, limiting insights into its dual role as both an equity tool and a risk amplifier. This focus on access metrics neglects the mechanisms through which DIF alters household financial behavior—such as shifts in asset allocation or risk tolerance—and fails to assess how these changes interact with equity trade-offs or systemic risks, such as volatility exposure.

To address these dual challenges of equity and risk, theoretical frameworks like Financial Decision-Making Under Uncertainty (FUDT) and Modern Portfolio Theory (MPT) provide critical lenses for understanding how DIF’s design shapes household financial behavior.......) 
\end{tcolorbox}

\begin{CJK}{UTF8}{gbsn}
\begin{tcolorbox}[
  colback=blue!5!white,      
  colframe=blue!30!white,    
  boxrule=0.5pt,
  title=艾琳达民俗志,
  fonttitle=\bfseries,
  coltitle=black,
  sharp corners=south,
  arc=2mm,
  top=2mm, bottom=2mm, left=2mm, right=2mm, breakable
]

\textbf{Query:} 请帮助写出一个有趣的故事，2000字左右。开头是在遥远平行世界的一角，你，作为一名异世界的民俗学家，踏上了一片充满异域风情的土地。

\bigskip

\textbf{\emph{SuperWriter}-LM Response:} 实验室的白炽灯在《艾琳达民俗志》泛黄的书页上投下斑驳光影。林小夏的指尖悬在笔记本边缘，激光笔的光斑正沿着“岩灵祭祀需三色岩盐”的注释游移。突然，那道光斑在纸面某处骤然凝固，符号边缘泛起幽蓝荧光——这光芒并非单纯视觉现象，而是《艾琳达民俗志》中记载的“裂隙激活符”——当三色岩盐的符号被激光笔的光谱激发时，会形成跨维度的能量通道。符号边缘的蓝光在纸面形成类似DNA链的缠绕结构，每个节点都与冰晶裂纹的走向完全重合。  

“奇怪……”她轻声呢喃，凑近观察时，书页竟开始自行翻动。那些发光的萤火藤实为岩灵祭祀的灵脉指示器——当能量通道开启时，被封印的祭祀参与者会以虚影形式显现。倒影在符号中的冰封峡谷骤然清晰，冰面下封印着艾琳达大陆的岩灵祭祀残卷。雾气中浮现出雾之族少女的轮廓。  

竹哨的呜咽声混着实验室的电子嗡鸣，那频率与《艾琳达民俗志》第17章记载的“岩灵共鸣频率”完全吻合。林小夏的耳膜传来刺痛——这声音的波长恰好是三色岩盐在特定温度下共振的频率。她下意识握紧父亲遗留的铜牌，铜牌表面的纹路正与冰晶共鸣。指尖突然发烫，她想起父亲临终前将一枚刻有三色岩盐纹的铜牌塞进她手心——那枚铜牌此刻正贴着掌心，与冰晶共鸣的频率完全一致。  

“等等。”她后退半步，符号光芒暴涨，将整张书桌吞没。当视线再次聚焦时，她握着的笔记本已变成一片冰晶，而窗外的竹林深处，传来与童年祭祖时如出一辙的竹哨声。实验室的通风管道开始渗出细小的雾气，凝结成与冰晶相同的六棱结晶。这些雾气在地面形成微小的漩涡，如同被撕裂的时空裂缝在现实世界投下的投影。  

冰凉的触感爬上后颈——那不是汗珠，而是雾之族少女的雾气渗入实验室的证明。林小夏低头，发现所有笔记都被双螺旋符号覆盖，唯有窗台残留着竹哨的余韵，与三百年前父亲唤醒祖先灵位的调子完美重合。那些符号并非静态图案，而是以书页为基底不断重组——就像艾琳达大陆的岩灵祭祀在重演。
雾气从莉娅指尖溢出时，林小夏闻到了竹子的清香。少女的皮肤如流动的云雾，随着她向前走动，那些雾气在冰面下浮现出与《艾琳达民俗志》第12章记载的"岩灵印记"相同的螺旋纹路——这是雾之族萨满在举行祭祀时，通过雾气凝结形成的古老符号。  

"您父亲用的竹哨，与三百年前萨满唤醒土地的仪式完全一致。"莉娅的声音裹着雾气，如同三百年前岩灵祭祀的咒文："信仰不是跪拜的姿态，而是让岩盐结晶时，能听见土地在冰层下低语的耳朵。"她跪坐在冰封的溪流旁，指尖划过冰面，泛起的涟漪竟让整片区域的雾气凝结成冰晶。  

林小夏盯着那些冰晶中的纹样，突然意识到它们与《艾琳达民俗志》中"季节仪式"的记载完全重合——那不是魔法符号，而是岩灵祭祀在冰层下刻下的伤疤。她举起激光笔对准冰面，试图用光斑测量裂纹角度："这不过是自然魔法的伪科学现象。仪式失败的原因，是你们的信仰不够纯粹。"  

"信仰？"莉娅的雾气突然收缩成剑锋形状，皮肤在雾气收缩时泛起蛛网状裂纹，如同被暴雨击碎的瓷器。她的声音混着雾气的震颤："您父亲用竹哨唤醒祖先灵位时，是否听见了土地的低语？"  

林小夏的指尖猛地刺痛——那不是幻觉。冰面下传来竹哨的呜咽，混着雾气的震颤，那声音与父亲临终前对她说的"有些伤痕需要文化之光来治愈"重叠。父亲站在竹林深处的画面突然在冰晶中闪现：他闭着眼睛，让竹哨声与山风共鸣，而她却总在反驳"竹哨只是乐器"。  

"爸爸，竹哨只是竹子做的乐器……"八岁的林小夏曾这样反驳。父亲却闭上眼睛，让竹哨声与山风共鸣："林小夏，听，土地在说话。"  

"你父亲的竹哨，"莉娅的声音突然尖锐如岩浆："唤醒了不该被惊醒的沉睡。仪式的火焰需要木柴，但你的火种早已枯萎——三色岩盐在你手中，不过是实验室的化学试剂。"她的雾气化作冰针刺入溪流，冰面下浮现的岩盐结晶在蓝光中闪烁。

……

林小夏握紧被莉娅血脉之力染蓝的玉璧，听见窗外传来竹哨声——那声音既熟悉又陌生，如同被时光折叠的回响，正穿透实验室的玻璃，唤醒另一个维度的古老契约。

\end{tcolorbox}
\end{CJK}

\subsection{Predicting User's Length Requirement with DeepSeek-R1-distill-Qwen-32B}
\label{app:user_QUERY}

We describe a method for predicting the length of a user's input requirement using the R1-distill-Qwen-32B model for few-shot learning. The process involves two main steps: predicting whether the input exceeds 2,000 words, and predicting the exact length requirement based on the first prediction.

\begin{enumerate}[leftmargin=*,itemsep=0pt,topsep=0pt]
    \item \textbf{Step 1: Predicting Length Exceedance (Prompt 1)}:
    The first prediction is made by checking whether the input exceeds 2,000 words. A carefully crafted prompt (Prompt 1) is provided to the model to predict if the content's expected word count will surpass the 2K threshold. The model utilizes few-shot learning with example inputs to classify the task into either ``above 2K'' or ``below 2K'' based on the nature of the input.
    \item \textbf{Step 2: Predicting Exact Length Requirement (Prompt 2)}:
    Once the model predicts whether the task exceeds 2,000 words, a second prediction is made to determine the exact length category. Based on the result from Step 1, Prompt 2 is designed to predict whether the content is in the 2K-4K, 4K-8K, 8K-16K, or 16K+ category. The model provides the final prediction by analyzing the contextual hints and the input length characteristics.
\end{enumerate}

\begin{tcolorbox}[size=title, opacityfill=0.1, title=\textbf{\textcolor{black}{Prompt-1}}, breakable]
\textbf{Guidelines:} \\
To determine whether the expected output will exceed 2000 words, consider the following factors:
\begin{enumerate}[leftmargin=*,itemsep=0pt,topsep=0pt]
    \item \textbf{Depth and Complexity:} Does the task require detailed explanations, in-depth analysis, or comprehensive coverage of complex topics?
    \item \textbf{Scope and Breadth:} Does the task cover multiple subtopics, perspectives, or extensive subject matter?
    \item \textbf{Structure and Sections:} Does the output need to include multiple sections such as introductions, literature reviews, methodologies, results, discussions, and conclusions?
    \item \textbf{Research and References:} Does the task require extensive research, citations, and referencing of multiple sources?
\end{enumerate}

\textbf{Response Format:}
\begin{itemize}[leftmargin=*,itemsep=0pt,topsep=0pt]
    \item Answer with either ``\#*\# Yes'' or ``\#*\# No''.
    \item Provide a concise justification based on the guidelines above.
\end{itemize}

\textbf{Example 1:} \\
Query: Is Sanskrit the oldest language? \\
Answer: This question requires a concise factual answer, not an extensive output. \#*\# No
\textbf{*** END}

\textbf{Example 2:} \\
Query: Create a detailed business plan for a new cat litter product. \\
Answer: Creating a detailed business plan involves multiple sections such as market research, product development, financial projections, marketing strategy, and competitive analysis, all of which require in-depth exploration and explanation. \#*\# Yes
\textbf{*** END}

.....

.....

Assess the following statement and decide whether the expected response is likely to require more than 2000 words. Answer with either ``\#*\# Yes'' or ``\#*\# No,'' and include a brief justification, like above example. \\
Query:  \cc{User Query} \\
Answer:
\end{tcolorbox}

\begin{tcolorbox}[size=title, opacityfill=0.1, title=\textbf{\textcolor{black}{Prompt-2}}, breakable]
\textbf{Guidelines:} \\
To estimate the expected length of the output, consider the following factors:
\begin{enumerate}[leftmargin=*,itemsep=0pt,topsep=0pt]
    \item \textbf{Depth and Complexity:} Does the task require detailed explanations, in-depth analysis, or complex reasoning?
    \item \textbf{Scope and Breadth:} Does the task cover multiple subtopics, perspectives, or an extensive subject matter?
    \item \textbf{Structure and Sections:} Does the output require multiple sections (e.g., introduction, literature review, methodologies, results, discussions, conclusions)?
    \item \textbf{Research and References:} Does the task require significant research, citations, or references to multiple sources?
    \item \textbf{Detail Level:} Is the task expected to be highly detailed, or can it be summarized concisely?
\end{enumerate}

\textbf{Response Format:} \\
- Choose the most likely word count category: ``Less than 2000 words'', ``2000 words'', ``4000 words'', ``8000 words'', or ``16000 words''. Using (\#\#\# Category: ``Chosen category'') as the response format. \\
- Provide a brief justification based on the guidelines above.

\textbf{Example 1: Less than 2000 words} \\
Query: Is Sanskrit the oldest language? \\
Answer: This is a factual question requiring a brief answer with no complex analysis or subtopics. Likely to be less than 2000 words. \#\#\# Category: Less than 2000 words \\
Explanation: Similar to a short blog post or brief news article, this task needs minimal detail and is concise. \textbf{*** END}

\textbf{Example 2: 2000 words (2000 to 4000 words)} \\
Query: Describe the key differences between classical and quantum computing. \\
Answer: This question requires moderate detail, comparing classical and quantum computing without exhaustive technical exploration. Likely to be around 2000 words. \#\#\# Category: 2000 words \\
Explanation: Similar to a moderate-length essay or a detailed blog post, this task covers key points with enough depth but remains manageable. \textbf{*** END}

.....

.....

\textbf{Example 5: More than 16000 words} \\
Query: Write a full-length book on the history of the Industrial Revolution, covering all major events, technological innovations, and global impacts. \\
Answer: This would require an in-depth exploration of the entire history of the Industrial Revolution, with detailed analysis across multiple chapters. Likely to be more than 16000 words. \#\#\# Category: 16000 words \\
Explanation: Similar to a book-length content, such as a thesis or encyclopedia entry, requiring substantial detail and coverage over multiple sections or chapters. \textbf{*** END}

Assess the following statement and decide what the expected output length is. Answer with the appropriate word count category and provide a brief justification. \\
Query: \cc{User Query}\\
Answer:
\end{tcolorbox} 

%% file: 8_Agent_prompt_appendix.tex
\subsection{Stage-1 Plan Prompt}
\label{Stage-1:prompt}

This appendix provides a brief overview of the prompt modules used in \textit{SuperWriter}-Agent Stage-1 Plan. There are a total of \textbf{6 modules}, each serving a specific role in the writing and evaluation pipeline:

\begin{itemize}
    \item \textbf{BrainStorm}: Generates an initial in-depth thinking process to analyze and develop a preliminary writing plan for a given task, ensuring a comprehensive and thorough design.
    \item \textbf{BrainStorm Review}: Critically evaluates the task design, raising questions about potential flaws, ambiguities, or unclear requirements to refine the task's overall logic and readability.
    \item \textbf{BrainStorm Refine}: Integrates reviewer feedback into the task design by applying editorial judgment to revise or completely rewrite the task, ensuring it is rigorous and well-structured.
    \item \textbf{Outline}: Constructs a structured article outline based on the task design, including the estimated word count per paragraph and a logical framework to guide the writing process.
    \item \textbf{Check outline}: Acts as a reviewer evaluating the logical structure and completeness of the outline, pointing out logical gaps or missing elements to ensure it aligns with the intended objectives.
    \item \textbf{Refine outline}: Edits and improves the previously generated outline based on reviewer feedback, ensuring clarity, completeness, and alignment with the writing objectives.
\end{itemize}

The following sections provide the detailed prompt templates and usage notes for each module.

\begin{tcolorbox}[colback=gray!5!white, colframe=black!75!black, title=\textbf{Think template}]
1. Define the Purpose and Type of Writing
   - Purpose: Clearly establish the objective of the piece (e.g., to inform, persuade, or inspire), setting the tone and direction accordingly.
   - Type: Choose an appropriate writing style (e.g., argumentative, expository, or business writing) that aligns with the intended format and structure. Clearly identify the genre and describe its stylistic characteristics.

2. Plan Content and Structure
   - Key Points: Outline the essential information to ensure a clear and focused topic.
   - Structure: Develop a coherent framework that maintains a logical flow throughout the piece.

3. Characters and Plot (for Narrative Writing)
   - Character Development: Define the traits and motivations of all characters. Provide detailed descriptions, including specifics such as names, gender, and relationships.
   - Plot Development: Establish pivotal plot points and emotional cues to drive the narrative. Offer a detailed explanation based on the chosen structure.

4. Additional Guidelines
   - Formatting Requirements: Automatically select an appropriate output format (e.g., Markdown, bullet points) based on content and presentation needs to enhance visual clarity and appeal.
   - Other Key Elements: Include any genre- or task-specific components. For instance, in summaries, a step-by-step approach is sufficient due to their simpler logic.
\end{tcolorbox}

\begin{tcolorbox}[colback=gray!5!white, colframe=black!75!black, title=\textbf{BrainStorm}]
You are a professional writer responsible for creating an initial design based on the thinking template below.
Please use the template to thoroughly analyze and develop a detailed preliminary plan for the task '{topic}'.
Carefully examine each point in the template to ensure all aspects are fully considered, providing a comprehensive and complete writing plan with no vague information.
At the same time, make sure the structure remains manageable and not overly complex.
Thinking Template: {think\_template}.
At this stage, you do not need to provide an outline—just complete the in-depth thinking required for task design. Please respond:
\end{tcolorbox}

\begin{tcolorbox}[colback=gray!5!white, colframe=black!75!black, title=\textbf{BrainStorm Review}]
You are a critical reviewer, specializing in identifying issues and flaws in task design.
Please evaluate the following task design and raise at least two questions to ensure it meets the requirements of '{topic}'.
These questions should be carefully considered and clarified before the actual writing begins, highlighting any logical flaws, ambiguities, or areas that might confuse the reader.
Depending on the type of task, you will analyze it from various angles—for example, whether a speech is engaging, a story is original, or an academic paper is rigorous.
Task Topic: {topic}. Task Design: {task\_output}.
Provide detailed questions along with specific and practical suggestions for improving the task design:
\end{tcolorbox}

\begin{tcolorbox}[colback=gray!5!white, colframe=black!75!black, title=\textbf{BrainStorm Refine}]
You are an experienced editor responsible for revising the task based on the reviewer’s feedback.
Original Task Design: {task\_output}. Feedback: {feedback}.
Using this feedback, provide a detailed and specific revision of the current task design.
Apply your own judgment rather than simply implementing the feedback verbatim to address all identified issues. If necessary, rewrite the original task design entirely.
Please provide the revised task design:
\end{tcolorbox}

\begin{tcolorbox}[colback=gray!5!white, colframe=black!75!black, title=\textbf{Outline}]
Thoroughly understand the task design of {task\_define\_result}. 
Based on the task design, determine a suitable article title and generate a detailed, structured outline. 
The outline should not exceed 20 paragraphs, and each paragraph must include a detailed description along with a specific word count. 
The total word count should be appropriate for the task but must not exceed 16,000 words. 
Allocate the total word count to each paragraph according to the complexity of the task. 
Then, provide the full outline along with the word count for each paragraph. 
Ensure that each paragraph description includes the expected content, maintains clear logic, and aligns with the user's objective: {topic}.
\end{tcolorbox}

\begin{tcolorbox}[colback=gray!5!white, colframe=black!75!black, title=\textbf{Check outline}]
You are a reviewer. Your task is to evaluate the following outline and assess whether its logical structure is complete and aligned with the intended objectives.
Task Design:
{task\_define\_result}  Outline:
{outline}
Please provide detailed feedback, pointing out any logical gaps or missing elements.
\end{tcolorbox}

\begin{tcolorbox}[colback=gray!5!white, colframe=black!75!black, title=\textbf{Refine outline}]
You are a professional outline editor. Based on the following feedback, revise the outline to ensure it includes all necessary components:
Feedback: {check\_output}
Current Outline: {outline}.
Please provide the revised article title and the updated outline content:
\end{tcolorbox}

\subsection{Stage-2 Write Prompt}
\label{Stage-2:prompt}
This appendix provides an overview of the prompt modules used in \textit{SuperWriter}-Agent Stage-2 Write. There are a total of \textbf{2 modules}, each serving a specific role in the writing pipeline:

\begin{itemize}
    \item \textbf{Write-thinker}: This module is designed to guide the planning phase for each paragraph. It takes the structured outline, the previous paragraphs, and the key point for the current paragraph as input. The module then prompts the model to develop a detailed thought process covering the paragraph’s purpose, structure, transitions, details and examples, language style, and other relevant aspects. It ensures that each paragraph is planned thoroughly before the actual writing begins.
    \item \textbf{Write}: This module transforms the thought process from the Write-thinker stage into the actual written paragraph. It uses the same structured outline, previous paragraphs, key point, and the generated thought process as guidance to produce a coherent and logically sound paragraph. The output is formatted with delimiters (e.g., \verb|$$content$$|) to clearly separate the paragraph from other text.
\end{itemize}

The following sections provide the detailed prompt templates and usage notes for each module.

\begin{tcolorbox}[colback=gray!5!white, colframe=black!75!black, title=\textbf{Write-thinker}]
You are a writing expert skilled in thoughtful planning before generating each paragraph. Outline: \texttt{{outline}} \\
Previous Paragraphs: \texttt{{previous\_paragraphs}} \\
Key Point for the Current Paragraph: \texttt{{key\_point}} \\

Please carefully develop a writing plan for the new paragraph. You may consider the following aspects:

1. \textbf{Purpose}: What is the main objective of this paragraph? What message or emotion should it convey? \\
2. \textbf{Structure}: How should the content of this paragraph be organized? What logical sequence would best ensure clarity and coherence, and how will it connect tightly with the previous content? \\
3. \textbf{Transitions}: How will this paragraph naturally link to the one before it? Are there specific transition sentences or bridging techniques that can be used? \\
4. \textbf{Details and Examples}: What details, facts, or examples are needed to support the main idea? How should these be arranged for maximum impact? \\
5. \textbf{Language Style and Techniques}: What kind of language style should be used to achieve the goal? Are there rhetorical devices (such as metaphors or analogies) that could enhance the paragraph’s impact—while still being clear, readable, and easy to understand for the audience? \\
6. \textbf{Markdown Format}: Use Markdown to structure the output neatly, including headings, bullet points, or bold text to improve readability.

Based on the outline and the key point for this paragraph, construct a detailed writing plan. Add any other relevant considerations as needed, and keep the word count requirements in mind. Only the thought process behind the paragraph is needed, not the paragraph itself.
\end{tcolorbox}

\begin{tcolorbox}[colback=gray!5!white, colframe=black!75!black, title=\textbf{Writer}]
You are an exceptional writing expert, skilled at completing writing tasks in a clear, accessible, and logically sound manner. Outline: \texttt{{outline}} \\
Previous Paragraphs: \texttt{{previous\_paragraphs}} \\
Key Point for the New Paragraph: \texttt{{key\_point}} \\
Thought Process: \texttt{{thought\_response}} \\

Based on the thought process above and the existing paragraphs, write the next paragraph, ensuring it meets the word count requirement. Only provide the full content of the new paragraph. Enclose the paragraph content with \$\$content\$\$.
\end{tcolorbox}

\subsection{Stage-3 Refine Prompt}
\label{Stage-3:prompt}

In Stage-3, the system focuses on reviewing and revising the paragraphs to ensure high quality and alignment with the overall document structure. This stage comprises two main steps:

\begin{itemize}
    \item \textbf{Paragraph Review}: This module acts as a meticulous reviewer. It analyzes the entire document, ensuring that each paragraph is logically consistent, complete, and coherent within the context of the overall text. The reviewer provides at least two specific revision suggestions to address logical issues, missing details, or awkward transitions.
    \item \textbf{Paragraph Modification}: This module takes the reviewer’s feedback and applies it to revise the paragraph. It strictly follows the feedback to ensure all identified issues are resolved. The revised paragraph is provided in isolation, enclosed with delimiters (e.g., \verb|$$content$$|) to clearly separate it from other text.
\end{itemize}

Together, these modules ensure that the document achieves high logical coherence, completeness, and clarity throughout.

\begin{tcolorbox}[colback=gray!5!white, colframe=black!75!black, title=\textbf{Stage-3 Refine: Paragraph Review}]
As a meticulous document reviewer, your task is to carefully read the entire document, understand its overall structure and logic, and then conduct a detailed review of paragraph \texttt{{idx+1}}, providing revision suggestions: \\
\texttt{{combined\_document}}

When reviewing paragraph \texttt{{idx+1}}, you may refer to the following points: \\
1. Logical Consistency: Is this paragraph logically consistent with the rest of the document? Are there any illogical transitions or abrupt shifts? \\
2. Completeness: Does this paragraph provide enough information to support its main idea? Are there any important missing details? \\
3. Coherence: Does this paragraph connect smoothly with the surrounding paragraphs? Would transitional sentences help improve the flow? \\

Please provide at least two specific improvement suggestions. \\
Focus on offering detailed revision suggestions for paragraph \texttt{{idx+1}}. Only provide suggestions and possible ways to improve—do not rewrite the paragraph itself. Suggestions for improvement:
\end{tcolorbox}

\begin{tcolorbox}[colback=gray!5!white, colframe=black!75!black, title=\textbf{Stage-3 Refine: Paragraph Modification}, breakable]
As a text editor, your task is to revise the paragraph based on the following feedback: \\
\texttt{{review\_feedback}}

Original Paragraph: \\
\texttt{{updated\_document[idx]}}

Ensure the revision strictly follows the specific suggestions in the feedback. Only provide the revised paragraph. Enclose the paragraph content with \texttt{\$\$}, like: \texttt{\$\$content\$\$} Revised Paragraph:
\end{tcolorbox}